%% file: main.tex
\documentclass[letterpaper, 10 pt, conference]{ieeeconf}  

\IEEEoverridecommandlockouts                              

\usepackage[whole]{bxcjkjatype} 
\usepackage{graphics} 
\usepackage{epsfig} 
\usepackage{mathptmx} 
\usepackage{times} 
\usepackage{amsmath} 
\usepackage{amssymb}  

\usepackage{multirow}
\usepackage[table,xcdraw]{xcolor}
\usepackage{threeparttable}
\usepackage{eucal}
\usepackage{balance}
\usepackage{mathrsfs}  
\usepackage{comment}
\usepackage{algorithm}
\usepackage[noend]{algpseudocode}
\usepackage{float}
\usepackage{textcomp}
\usepackage{mathcomp}
\usepackage{bm}
\usepackage{url}
\usepackage{comment}
\usepackage{booktabs}
\usepackage{cite}
\def\eg{\textit{e.g}\onedot} 
\def\ie{\textit{i.e}\onedot}

\def\etal{\textit{et al}\onedot}

\makeatletter
\let\MYcaption\@makecaption
\makeatother

\usepackage[font=footnotesize]{subcaption}

\makeatletter
\let\@makecaption\MYcaption
\makeatother

\usepackage{subcaption}

\newcommand{\honda}[1]{#1}

\newcommand{\DIFF}[1]{\textcolor{orange}{#1}}

\newcommand{\myvec}[1]{\mathbf{#1}}
\newcommand{\myarray}[1]{\textbf{\textit{#1}}}
\newcommand{\myset}[1]{\mathcal{#1}}

\newcommand{\kozuno}[1]{\textcolor{magenta}{#1}}
\usepackage[colorinlistoftodos]{todonotes}

\def\eg{{\it e.g.}}

\def\ie{{\it i.e.}}
\def\etal{{\it et al. }}

\def\methodname{DRL replanner}
\def\methodlong{replanning controller}

\title{\LARGE \bf
When to Replan? An Adaptive Replanning Strategy\\ for Autonomous Navigation using Deep Reinforcement Learning
}

\author{Kohei Honda$^{1}$, Ryo Yonetani$^{2}$, Mai Nishimura$^{2}$, and Tadashi Kozuno$^{2}$
\thanks{$^{1}$Kohei Honda is with the Department of Mechanical Systems Engineering, Nagoya University, Furo-cho, Chikusa-ku, Nagoya, Aichi, Japan. This work was done while he was a research intern at OMRON SINIC X Corporation. {\tt\small honda.kohei.b0@s.mail.nagoya-u.ac.jp}}%
\thanks{$^{2}$Ryo Yonetani, Mai Nishimura, and Tadashi Kozuno are with OMRON SINIC X Corporation, Hongo, Bunkyo-ku, Tokyo, Japan. {\tt\small \{ryo.yonetani, mai.nishimura, tadashi.kozuno\}@sinicx.com}}%
}

\begin{document}

\maketitle
\thispagestyle{empty}
\pagestyle{empty}

\begin{abstract}

\input{src/abstract.tex}

\end{abstract}


\input{src/introduction.tex}

\input{src/related_work.tex}

\input{src/hierachical_planning_framework}

\input{src/adaptive_replanning}

\input{src/experiments}

\input{src/conclusion.tex}







\bibliographystyle{./IEEEtran}
\bibliography{./IEEEabrv, ./reference}

\end{document}

%% file: src/abstract.tex
The hierarchy of global and local planners is one of the most commonly utilized system designs in autonomous robot navigation. While the global planner generates a reference path from the current to goal locations based on the pre-built map, the local planner produces a kinodynamic trajectory to follow the reference path while avoiding perceived obstacles. To account for unforeseen or dynamic obstacles not present on the pre-built map, ``when to replan'' the reference path is critical for the success of safe and efficient navigation. However, determining the ideal timing to execute replanning in such partially unknown environments still remains an open question. In this work, we first conduct an extensive simulation experiment to compare several common replanning strategies and confirm that effective strategies are highly dependent on the environment as well as the global and local planners. Based on this insight, we then derive a new adaptive replanning strategy based on deep reinforcement learning, which can learn from experience to decide appropriate replanning timings in the given environment and planning setups. Our experimental results show that the proposed replanner can perform on par or even better than the current best-performing strategies in multiple situations regarding navigation robustness and efficiency.

%% file: src/introduction.tex
\section{INTRODUCTION}
\label{sec:introduction}
\begin{figure*}[t]
  \begin{minipage}[t]{0.4\linewidth}
    \centering
    \includegraphics[keepaspectratio, scale=0.45]{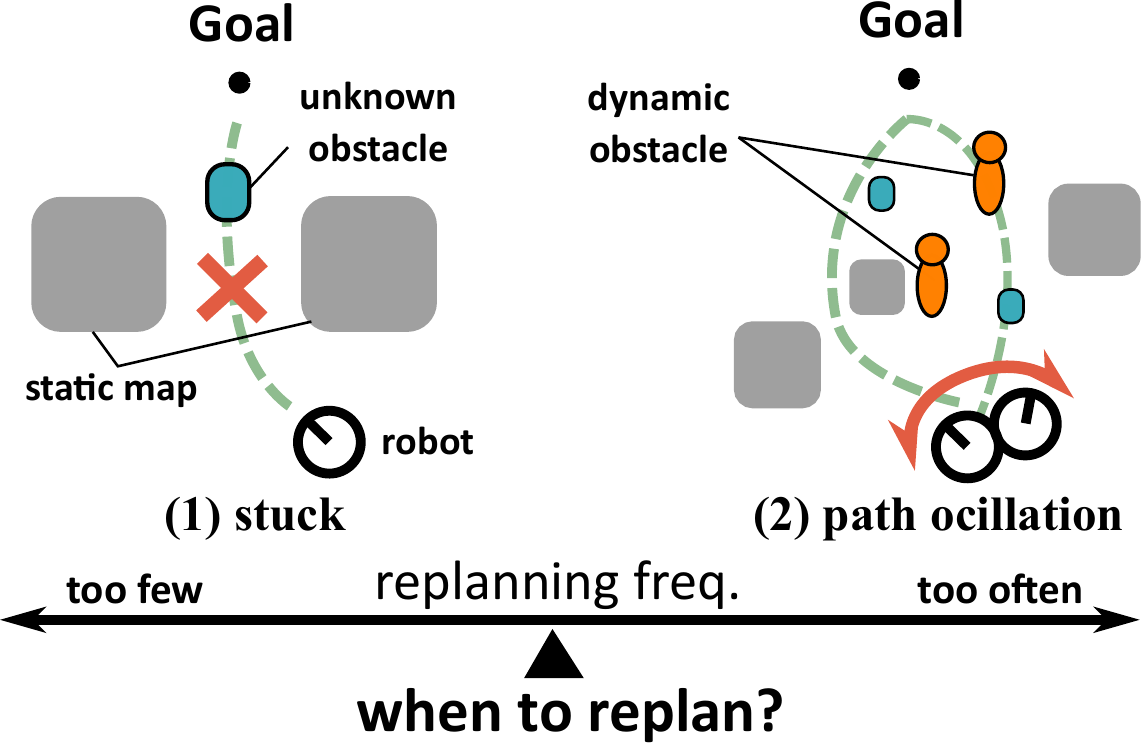}
    \subcaption{Learning when to replan}
    \label{fig:stuck_and_oscillation}
  \end{minipage}
  \begin{minipage}[t]{0.6\linewidth}
    \centering
    \includegraphics[keepaspectratio, scale=0.45]{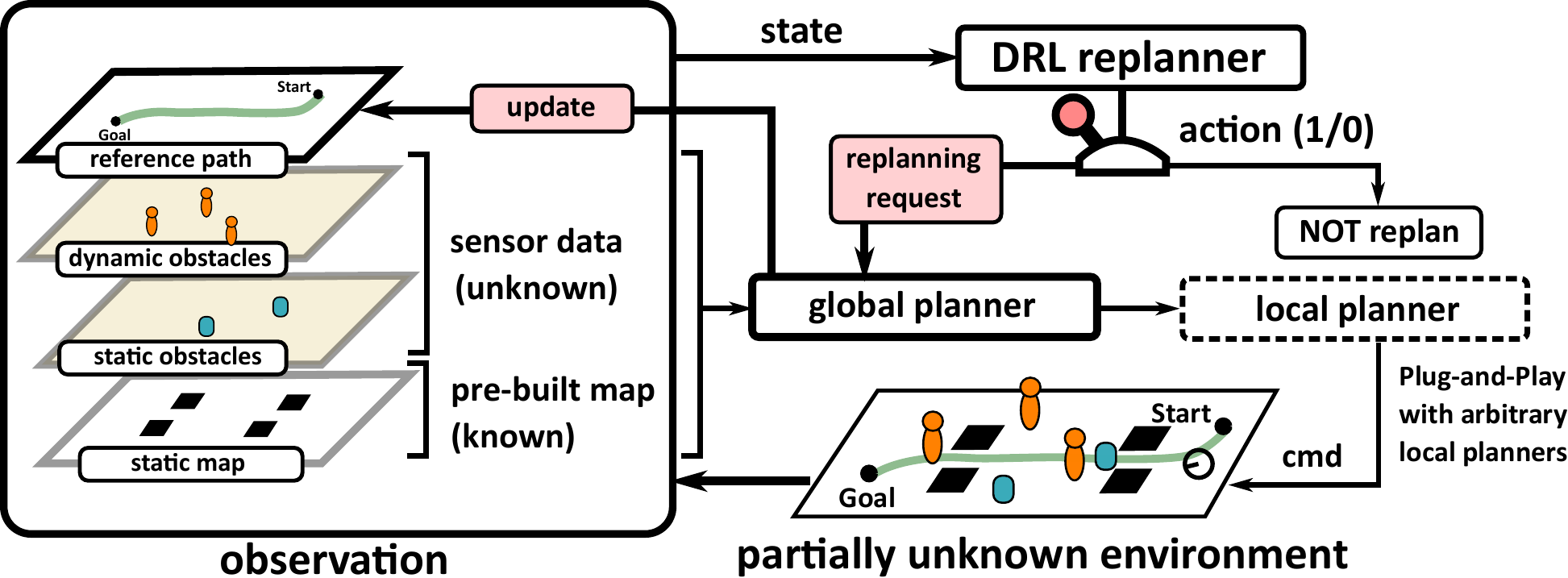}
    \subcaption{Proposed system overview}
    \label{fig:system_overview}
  \end{minipage}
  \vspace{-3mm}
  \caption{Overview of our work. The hierarchical planning framework can partition difficulties between global and local planning. However, depending on the timing of the global planner's replanning, negative behaviors can occur (Fig.~\ref{fig:stuck_and_oscillation}). Specifically, in partially uncharted terrains that include unforeseen and dynamic obstacles on the pre-built static map, insufficient replanning frequency causes negative behaviors, \ie, (1) too low a replanning frequency can cause the robot to get stuck, while (2) excessive replanning can lead to path oscillation. Our objective is to control the timing of replanning adaptively to enable robust and efficient navigation of AMRs. To achieve this, we propose a replanning controller based on DRL (Fig.~\ref{fig:system_overview}). The DRL replanner obtains an efficient and adaptive replanning strategy by training on the partially uncharted terrains and the hierarchical planning framework as its environment.}
\label{fig:crown_jewel}
\vspace{-6mm}
\end{figure*}

It's a fact of life that things do not always go as planned. Whether the unexpected is a minor setback or a major obstacle, we must be ready to pivot and adjust our plans to ensure that we can still achieve our goals. The same applies to navigating autonomous mobile robots (AMRs). In real-world scenarios such as industrial factories or busy restaurants, the environment is filled with unforeseen obstacles or pedestrians that diverge from the pre-built map. To deal with such partially uncharted terrain, it becomes imperative to dynamically replan the pre-planned paths as needed.

In this paper, we delve into the \emph{timing} of the replanning feature in the common hierarchical planning framework~\cite{brock1999high}.
Carefully tuning the replanning feature is crucial in practice, as it can drastically change the behavior of AMRs and can affect navigation robustness and efficiency.
When performed at the right time, replanning can enable goal-oriented and reactive motion in the presence of unforeseen and dynamic obstacles. However, improper replanning, for example, if done too frequently or infrequently, can also cause the AMRs to perform inefficient travel (\eg, path oscillation) or even get completely stuck, as shown in Fig.~\ref{fig:stuck_and_oscillation}\footnote{See the demo video for more details: \url{https://www.youtube.com/watch?v=W8nBFKDxsb0}}. 
Despite some relevant work~\cite{murphy1997explicitly, tordesillas2019real, zha1997detecting, oleynikova2018safe}, an effective replanning strategy -- more specifically \emph{when to execute replanning for robust and efficient navigation in partially uncharted terrain} -- remains an open question.

\looseness=-1
The primary contribution of this work is three-fold. First, we conduct a comprehensive experiment with simulated environments and systematically evaluate various replanning strategies commonly used in ROS 2 Navigation Stack~\cite{nav_stack2}. We demonstrate that effective strategies are highly dependent on the map layouts as well as on the global and local planning algorithms, which implies the fact that the replanning strategy should be carefully designed and tuned for every single environment and choice of planning algorithm (Section~\ref{sec:experiment}).

Second, we formulate a task of controlling the replanning timings for a global planner as a sequential decision-making problem using a Partially Observable Markov Decision Process (POMDP)~\cite{kaelbling1998planning} (Section~\ref{sec:hierarchical_planning}).
We consider a replanning controller as a decision maker that determines whether to replan the reference path at every timestep. 
The conventional rule-based replanning strategies can be viewed as policies in the POMDP, which allows us to compare diverse replanning strategies in the unified framework.


Finally, based on the aforementioned POMDP, we derive a Deep Reinforcement Learning (DRL)-based replanning controller (hereafter referred to as \emph{DRL replanner}) that learns to decide when to execute replanning for improving navigation robustness and efficiency in the current situation, as shown in Fig.~\ref{fig:system_overview} (Section~\ref{sec:proposed_method}).
The DRL replanner is trained with a standard DRL algorithm such as a deep Q network~\cite{dqn_nature}.
Notably, the DRL replanner can act as a drop-in replacement for the rule-based replanning strategy in existing hierarchical planning frameworks.
Our extensive experimental results have demonstrated that the proposed DRL replanner can achieve navigation that is as robust and efficient as, or better than, the currently best-performing strategies across various combinations of global and local planners in floor environments (Section~\ref{sec:experiment}).
These results have strong implications that well-controlled timing of replanning has the potential to achieve more robust and efficient navigation on the existing hierarchical planning frameworks by learning the environment-specific adaptive replanning strategy.

%% file: src/related_work.tex
\section{RELATED WORK}
\label{sec:related_work}
While partitioning complex navigation problems into global and local planning can increase substitutability and reduce computational complexity through parallel processing, 
it can cause inefficient path replanning such as getting stuck and path oscillation, as illustrated in Fig.~\ref{fig:stuck_and_oscillation}.
Existing approaches to this issue can be broadly categorized into the following three types:

\subsubsection{Replanning Timing Strategy}
Existing planning systems typically employ rule-based replanning strategies that need to be hand-engineered to account for the characteristics of the environments and planners. 
Indeed, Murphy \etal found in their early study that the timing of replanning could be a salient factor for navigation performance~\cite{murphy1997explicitly}.
Since then, various systems that replan at regular time intervals have emerged \cite{brock1999high}, and others adopt event-based rules, such as deviation from the reference path and detecting stuck \cite{ota2020efficient, kastner2021connecting}.
A practical software framework, ROS 2 Navigation Stack~\cite{nav_stack2}, provides behavior tree-based tools to allow engineers to implement various replanning strategies. 
However, these manual design approaches require significant expertise to fine-tune, and it can be challenging to adjust replanning rules adaptively on the basis of the situation.
This paper focuses for the first time on the evaluation and improvement of navigation performance through replanning strategies, which have not been adequately explored.

\subsubsection{Reference Path Modification}
Another approach to reducing inefficient path replanning is to modify the reference path in accordance with the current situation. 
For example, Tordesillas \etal proposed modifying a part of the reference path when the global planner significantly changes it from the current one~\cite{tordesillas2019real}.
While path modification is a valid approach, it highly depends on the nature of the global and local planners because the method and timing of the modification need to consider the shape of the reference path and the tracking performance of the local planner. 
Other works have also proposed updating the map dynamically \cite{zha1997detecting, oleynikova2018safe}, but this is only effective when all obstacles remain static.

\subsubsection{Navigation based on DRL}
In recent years, DRL has become a popular approach for point-to-point navigation~\cite{dong2021review}. While many studies employ DRL to learn adaptive behavior for local planners in short-range navigation, some recent studies use DRL planners in conjunction with the reference paths generated by classical global planners in long-range navigation~\cite{prm-rl,rl-rrt,pokle2019deep}. 
Most of these works, however, assume a standard periodic replanning of the global planner~\cite{ota2020efficient, kastner2021connecting, kastner2021arena, angulo2022policy}.
An exception to this norm is the work by Wang \textit{et al.}, which does not necessitate replanning but is constrained to environments with a discrete action space~\cite{wang2020mobile}.
Our work fundamentally differs from these approaches in that we propose a DRL-based adaptive replanning strategy for the global planner.
Note that our approach is not limited to any particular type of local planner and can be used with a diverse array of planning methods. This adaptability is a key strength of our approach, allowing us to provide a flexible and versatile framework.

%% file: src/hierachical_planning_framework.tex
\section{HIERARCHICAL PLANNING FRAMEWORK}
\label{sec:hierarchical_planning}

Our aim is to optimize the timing of replanning for robust and efficient navigation of AMRs in partially uncharted terrain due to unforeseen or dynamic obstacles that were not present when building an environment map. The replanning timing is desirable to be adaptively determined based on the nature of the hierarchical planning framework and observations of the robot's surroundings obtained from sensors. we first provide an overview of a conventional hierarchical planning framework with a replanning feature and then formulate how to control the replanning timings as a sequential decision-making problem with a partially observable Markov decision process (POMDP)~\cite{kaelbling1998planning}.



\subsection{Global and Local Planners}
\label{subsec:hierachical_system}

The typical configuration of a hierarchical planning framework consists of asynchronously operating \emph{global planner} and \emph{local planner} modules as shown in Fig.~\ref{fig:system_overview}.
The global planner (represented by $f_{\rm{gp}}$) computes a reference path to a goal $\myvec{p}^g \in \mathbb{R}^2$ from the robot position based on observations from onboard sensors and \honda{a prebuilt map that contains information about known obstacles and no-entry areas.}

Specifically, $f_{\rm{gp}}$ receives the sensor observation and pre-built map information $\myset{M}_t$ and robot position $\myvec{p}_t \in \mathbb{R}^2$ at time $t$ and produces a reference path as the sequence of 2D positions $\myarray{P}^r_{t + \Delta t_{d}} = \{\myvec{p}^r_0, \myvec{p}^r_1, \dots\}$ for time $t + \Delta t_{d}$: $\myarray{P}^r_{t + \Delta t_{d}} = f_{\rm{gp}}(\myvec{p}^g, \myvec{p}_t , \myset{M}_t)$, 
where $\Delta t_{d}$ is the time delay due to the computation time for global planning.
This calculation can generally become expensive as the size of pre-built maps increases, and $\Delta t_d$ can be larger than the control interval $\Delta t$.
Note that $\Delta t_d$ can be estimated to some extent by using an any-time algorithm (\eg, any-time A*~\cite{likhachev2003ara}) because they are interruptible.
In practice, to consider the environmental change, global planning is performed repeatedly at a low frequency (\textasciitilde 1 Hz) based on a predefined replanning strategy.

The local planner $f_{\rm{lp}}$ generates a more fine-grained and collision-free control command with a given control frequency (10 Hz\textasciitilde) to guide the robot along the reference path.
The control command $\myvec{v}^{\rm{cmd}}_t$ at time $t$ is calculated from the the latest reference path $\myarray{P}_{t^*}$, the robot position $\myvec{p}_t$, and the sensor observation and pre-built map information $\myset{M}_t$ at time $t$, as $\myvec{v}^{\rm{cmd}}_t = f_{\rm{lp}} (\myarray{P}_{t^*}, \myvec{p}_t , \myset{M}_t)$,
where $t^*$ is delayed by at most $\Delta t_d$ from time $t$.
The control command $\myvec{v}^{\rm{cmd}}_t$ updates the physical motion of the robot for the control interval $\Delta t$.

\subsection{Formulation using POMDP}
\label{subsec:pomdp_definition}

\begin{figure}[t]
  \centering
  \includegraphics[width=\linewidth]{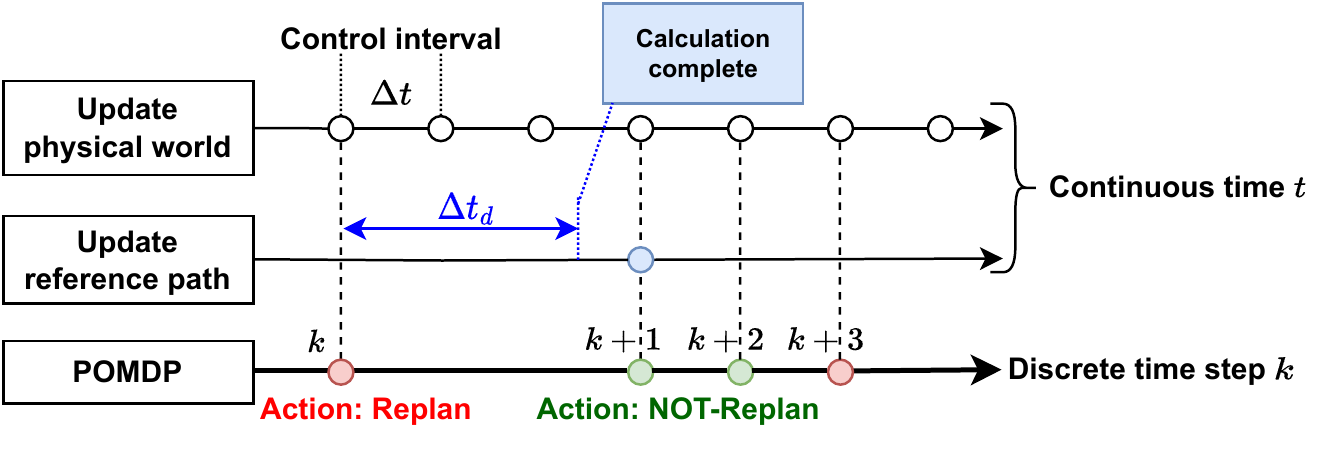}
  \vspace{-7mm}
 \caption{Time transition according to the replanning action.}
  \label{fig:time_step}
  \vspace{-7mm}
\end{figure}

The replanning of reference paths in the above hierarchical planning framework can be modeled by a deterministic POMDP as follows. 
Let $\Pi = (\myset{S}, \myset{A}, \myset{O}, T, R, \Omega,  \gamma)$ be the POMDP tuple consisting of state space $\myset{S}$, action space $\myset{A}$, observation space $\myset{O}$, transition function $T: \myset{S}\times \myset{A} \mapsto \myset{S}$, reward function $R: \myset{S} \times \myset{A} \mapsto \mathbb{R}$, observation model $\Omega: \myset{S} \mapsto \myset{O}$, and discount factor $\gamma \in [0, 1)$.

The state $s_k$ includes the robot's status (position, orientation, and past trajectory), the planners' status (goal position and reference path), and the situation regarding surrounding obstacles at discrete time step $k$.
While the states of the robot and planners are essentially observable, the surrounding situation is only partially observable from onboard sensors. 
As a result, the DRL replanner can only obtain observations $o_k \in \myset{O}$ based on the observation model $o_k = \Omega(s_k)$ described in Section \ref{subsec:observation}.
The action $a_k \in \myset{A}$ is a binary decision of whether to replan or
not at time step $k$. Specifically, $\myset{A} = \{a_{\rm{rep}}, a_{\rm{not}}\}$, where $a_{\rm{rep}}$ requests the global planner $f_{\rm{gp}}$ to compute a new reference path and $a_{\rm{not}}$ does not, as illustrated in Fig.~\ref{fig:system_overview}.

The state $s_k$ transitions to $s_{k+1}$ with the transition function $T(s_k, a_k)$.
As the global and local planners run asynchronously and there is a delay due to the computation time for the global planning to output a new reference path, the transition function $T(s_k, a_k)$ needs to consider them.
In the transition function, if the action is not replanning, the physical world is rolled out by the control command from the local planner with the current existing reference path at time step $k$ and returns the state after $\Delta t$, \ie, $T(s_k, a_{\rm{not}})$ returns the state at the next time step $k+1=t+\Delta t$.
If the action is replanning, the global planner immediately starts computing a new reference path at time step $k$. 
At the same time, the control command by the local planning rolls out the physical world based on the existing reference path until the computation is finished.
Then, after computing the time delay $\Delta t_d$, the global planner outputs a new reference path, the local planner outputs a control command using the updated reference path, and the physical world returns a new state after $\Delta t$, \ie, $T(s_k, a_{\rm{rep}})$ returns the state at $k+1=t + (\lfloor \frac{\Delta t_d}{\Delta t} \rfloor + 1) \Delta t$, where $\lfloor \; \rfloor$ is a floor function.
Overall, the interval of time steps in the POMDP is variable depending on the action, as shown in Fig. \ref{fig:time_step}.

\subsection{Existing Replanning Strategies}
\label{subsec:baselines}

In this work, we compare four types of rule-based replanning strategies available in ROS 2 Navigation Stack~\cite{nav_stack2}.
\begin{itemize}
    \item \textit{Distance-based} strategy determines replanning timings on the basis of traveled distance, \eg, every $d_{\rm{rep}}$ meters. That is,
    if $\Delta d_k \geq d_{\rm{rep}}$, then $a_k= a_{\rm{rep}}$, otherwise $a_k= a_{\rm{not}}$, where $\Delta d_k$ is the difference in travel from the last replanning and $d_{\rm{rep}}$ is a given parameter.
    \item \textit{Stuck-based} strategy decides to execute replans when the robot stops at the same position for a given $\Delta t_{\rm{stuck}}$ seconds because the robot is considered to be stuck.
    \item \textit{Time-based} strategy performs replanning at every fixed period of $\Delta t_{\rm{rep}}$ seconds. $\Delta t_{\rm{rep}}$ should be larger than $\Delta t_d$ to consider the computation time of global planning.
    \item \textit{Time-with-patience} strategy adopts the \textit{time-based} strategy when the robot is far from the goal ($> d_{\rm{patience}}$) and changes to \textit{stuck-based} otherwise ($\leq d_{\rm{patience}}$), expecting to prevent a large detour near the goal.
\end{itemize}
Specific parameter settings will be presented in the experiment section. A key point here is that all of these existing strategies can be viewed as an instance of a hand-designed, deterministic \emph{policy} for the aforementioned POMDP, which takes the current observation as input to decide if replanning should be done as an action.

%% file: src/adaptive_replanning.tex
\section{ADAPTIVE REPLANNING USING DRL}
\label{sec:proposed_method}
Although a variety of replanning strategies are available, it remains unclear which one should be used and how the parameters should be tuned for a given environment as well as the choices of global and local planners. While replanning regularly with time-based and distance-based strategies at the highest possible frequency can allow the robot to constantly track the shortest distance path, doing so becomes superfluous if there are not many unforeseen obstacles in the pre-built map. Too much replanning could also cause path oscillation, especially when sampling-based global planners are utilized or the environment has many branching pathways. Moreover, since the computational cost of global planning increases as the environment becomes larger, it is important to execute replanning only when necessary. Adopting a stuck-based strategy is nonetheless nontrivial, because what can be defined as \emph{getting stuck} will depend on the performance of local planners and the dynamics of surrounding obstacles, making it harder to manually tune the parameters of the strategy.
To this end, we explore the possibility of leveraging deep reinforcement learning for adapting a replanning strategy to a given environment as well as the choices of planners.

\subsection{DRL-based Replanning Controller}
We derive a \emph{\methodlong} that can learn from its previous navigation experiences to create a better replanning timing for navigation efficiency and robustness. As illustrated in Fig.~\ref{fig:system_overview}, the replanner's action is essentially the same as that of existing replanning strategies, \ie, binary actions indicating whether or not to execute replanning to produce a new reference path for the local planner after the current time step $k$. 
In other words, the replanner can potentially be utilized as a replacement module for the replanning strategy in existing planning frameworks, thus making it compatible with various combinations of planners and other modules.


\subsection{Observation Design}
\label{subsec:observation}
We define the observation model $\Omega$ to include the status of surrounding obstacles, the reference path, the past trajectory, and the position of the target goal at time $k$. These observations are provided after being transformed onto the robot coordinate system, where the robot's forward direction aligns with the x-axis, thereby implicitly incorporating the robot pose into the observations. Note that these observations are also easily accessible in the practical navigation system. Specifically, the replanner receives an observation $o_k=[\myarray{S}_k, \hat{\myarray{P}}^r_{k^*}, \myarray{T}_k, \myvec{p}^g_r]  = \Omega(s_k)$, where $\myarray{S}_k = \{\myvec{s}_i\}_{i=0}^{n_s} \in \mathbb{R}^{2 \times n_s}$ is the two-dimensional scan positions down-sampled to $n_s$ at time step $k$, $\hat{\myarray{P}}^r_{k^*} = \{\myvec{p}_i\}_{i=0}^{n_p} \in \mathbb{R}^{2 \times n_p}$ is the latest reference path down-sampled to $n_p$ at time step $k$, which is computed by the global planner, $\myarray{T}_k = \{\myvec{t}_i\}_{i=0}^{n_t} \in \mathbb{R}^{2 \times n_t}$ is the robot's past trajectory downsampled to $n_t$ at time step $k$, which it does to make the replanner aware of stack and path oscillation situations.
$\myvec{p}^g_r \in \mathbb{R}^2$ is the relative position of a given goal. That is, for our POMDP, $\myset{O} \in \mathbb{R}^{2 \times (n_o + n_p + n_t +1)}$. Specific parameter settings will be described in the experiment section.

\subsection{Reward Design}
\label{subsec:reward}
Designing an appropriate reward function is a crucial step to enable reinforcement learning. The reward function should reflect and quantify the replanner's objective, that is, the efficiency (\ie, quick goals) and robustness (\ie, safety, collision-free) of the navigation in our case. As a unified metric that involves these criteria, we borrow the idea of \emph{success-weighted by normalized goal time (SGT)} ~\cite{barn_challenge}.
In the SGT, the score $s_{\rm{sgt}}^i$ of an episode $i$ is defined as follows:
\begin{align}
    s_{\rm{sgt}}^i
    =
    \frac{
        \myvec{1}_{\rm{suc}}^i \rm{OT}_i
    }{
        \rm{{clip}(\rm{AT}_i, \alpha \rm{OT}_i \beta \rm{OT}_i)}
    }
    \text{, where }
    \rm{OT}_i
    =
    \frac{
        L_{\rm{path}}^i
    }{
        \rm{speed_{\rm{max}}}
    }, \label{eq:sgt}
\end{align}
and $\myvec{1}_{\rm{suc}}^i$ is a binary indicator function of success that the robot reaches the goal without collisions.
$\rm{AT}_i$ and $\rm{OT}_i$ denote the actual and optimal traversal time as an indicator of the difficulty of the environment, respectively.
The clip function clips AT within $\alpha$OT and $\beta$OT ($0 < \alpha < \beta$) to reduce the influence of extremely easy or difficult episodes.
$L_{\rm{path}}^i$ is the optimal (shortest) path length to the goal and is calculated using the observation at the initial time by the Dijkstra method. 
$\rm{speed}_{\rm{max}}$ is a maximal speed of the robot.
We can define the reward function as $R(s_t, a_t) = \myvec{1}_{\myvec{p}_t = \myvec{p}^g} s_{\rm{sgt}}^i$, 
where $\myvec{1}_{\myvec{p}_t = \myvec{p}^g}$ is a binary function indicating that the robot reaches the goal.
Episodes are terminated when the robot collides or reaches the goal.
Although the SGT in (\ref{eq:sgt}) returns a non-zero score when an episode ends by reaching a given time limit, the reward at the time limit is zero to avoid conflicts with the handling of time limits in the bootstrapping of training~\cite{pardo2018time}.

\subsection{Learning Algorithm}
\label{subsec:training_procedure}
To train the replanning controller, a DRL algorithm that can handle binary actions would be sufficient.
We opt to use the popular deep Q-network (DQN) \cite{dqn_nature} algorithm that approximates the Q-function with a deep neural network (Q-network) and obtains an optimal policy that maximizes the Q-values.

A key point here is that the actual timing at which replanning is necessary appears only sparsely during a long travel. In other words, most of the gathered experiences, \ie, transitions, may not necessarily be useful for learning appropriate replanning timings. To address this issue, we employ a prioritized experience replay (PER) buffer~\cite{schaul2016prioritized} that gives different weights for each experience in the loss function based on its priority. Specifically, we use the priority $p_k$ based on the difference of Q-values between whether or not replanning was performed,
\begin{align}
    & p_k = \| Q_\theta(s_k, a_{\rm{rep}}) - Q_\theta(s_k, a_{\rm{not}}) \|.
\end{align}
Here, higher differences indicate that replanning at the corresponding timing makes the SGT better (or worse), and thus should be emphasized more in the replay buffer. This definition of priority is more intuitive and effective than the conventional priority based on TD-error (\ie, $p_k = \| Q_\theta(s_k, a_k) - (R_i(s_k, a_k) + \gamma \max_a Q^{\tau}_{\theta^{-}}(s_{k+1}, a))\|$), as will be shown empirically in our experimental results.


%% file: src/experiments.tex
\section{EXPERIMENTS}
\label{sec:experiment}
We conduct a comprehensive simulation study to systematically evaluate the existing planning strategies presented in Sec.~\ref{subsec:baselines} and the DRL replanner proposed in Sec.~\ref{sec:proposed_method}.
\begin{figure*}[t]
  \begin{minipage}[t]{0.5\linewidth}
    \centering
    \includegraphics[keepaspectratio, scale=0.43]{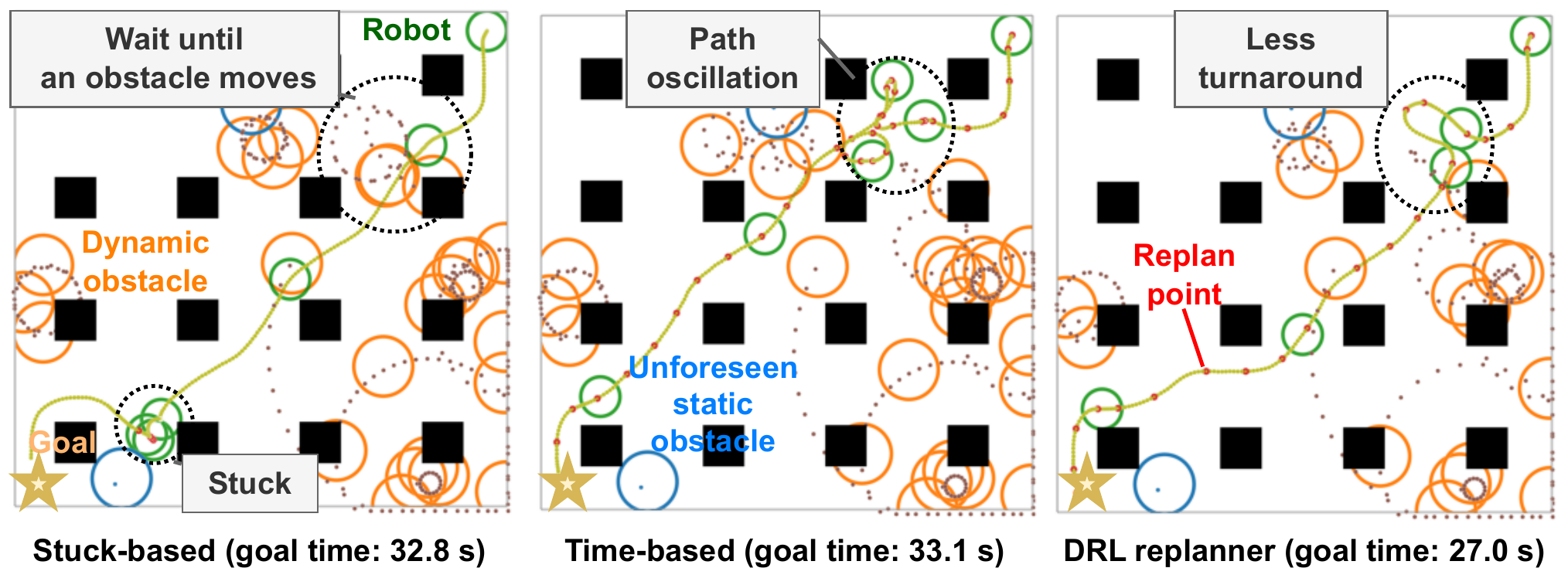}
    \subcaption{A simulation result for 16-pillar map}
    \label{fig:sim_result_16pillar}
  \end{minipage}
  \begin{minipage}[t]{0.5\linewidth}
    \centering
    \includegraphics[keepaspectratio, scale=0.43]{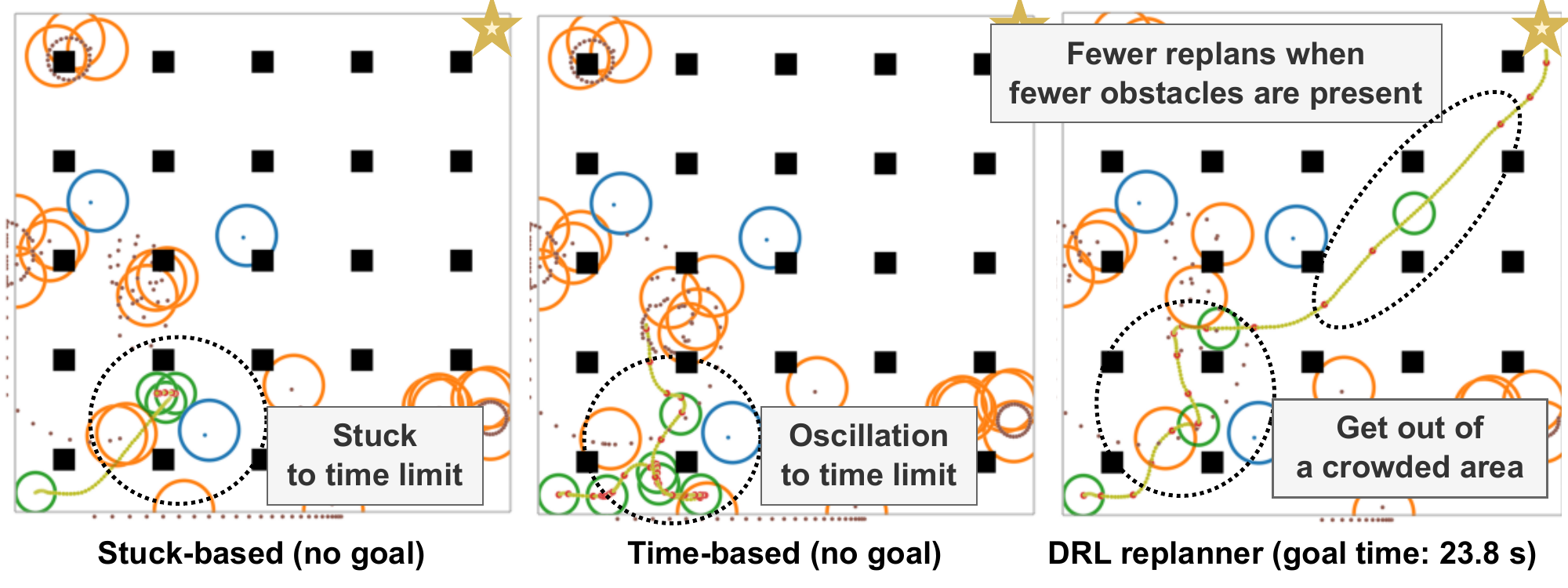}
    \subcaption{A simulation result for 25-pillar map}
    \label{fig:sim_result_25pillar}
  \end{minipage}
  \vspace{-3mm}
  \caption{Selected simulation results. In Fig. \ref{fig:sim_result_16pillar}, the stuck-based and time-based replanning strategies had difficulties in dealing with dynamic obstacles, while the proposed DRL replanner avoided them efficiently, resulting in reaching the goal about six seconds faster than the time-based replanning. In Fig. \ref{fig:sim_result_25pillar}, the stuck-based and time-based replanning strategies failed to reach the goal due to the movement of dynamic obstacles, while the proposed DRL replanner reached the goal by replanning at key points and adapting to the situation.}
\label{fig:simulation_result}
\vspace{-6mm}
\end{figure*}

\subsection{Environment Setup}
\label{subsec:environment}
We developed three navigation environments in continuous action space with different numbers of no-entry areas for the robot (nine, 16, and 25 uniformly lined regions sized 1.5, 1.0, and 0.5 meters, respectively; see Fig.~\ref{fig:simulation_result}). 
In each layout, dynamic obstacles and unforeseen static obstacles appears. The robot's objective is to reach a given goal while avoiding collisions with these obstacles and detouring around the no-entry areas. While the no-entry areas are encoded in the pre-built map used in the planners, the obstacles are not. To enable the dynamic obstacles to be widely distributed throughout the field, we assume that the dynamic obstacles can move over the no-entry areas.

The behavior of the dynamic obstacles is modeled as a social force model (SFM) \cite{social_force_model} or reactive stop model (RSM), which predicts its motion for several (\eg~3) seconds as a point mass model and stops in the case of a collision.
SFM simulates an agent that recognizes and avoids the robot, while RSM simulates an agent that does not avoid the robot. These models are selected for each obstacle and represent the dynamics that the obstacle may or may not yield to the robot. In these cases, where all dynamic obstacles give the robot the way based on the SFM, the robot does not need to replan and detour to reach the goal. However, such cases are rare in the real world.

\honda{
To evaluate how our approach generalizes to various obstacle settings, we spawn ten dynamic or unforeseen static obstacles with random initial positions, velocities, and dynamics models for each episode. The layout of the no-entry areas is fixed during both training and evaluation.
}

\subsection{Setup of Robot and Planners}
\label{subsec:implementation}
We simulate a circular wheeled robot (radius: 1.0 m). The robot follows the non-holonomic kinematics of the differential wheeled model with a maximum velocity (1.0 m/s) and angular velocity (1.0 rad/s) at each control interval $\Delta t = 0.1$ s. The initial and goal positions are sampled randomly from one of the corners of the environment.

As the hierarchical planning system described in Sec.~\ref{sec:hierarchical_planning}, we implemented Dijkstra, rapidly-exploring random tree* (RRT*)~\cite{karaman2011sampling}, and probabilistic roadmaps (PRM)~\cite{PRM} as the global planners, and implemented the dynamic window approach (DWA)~\cite{dwa} and sampling-based model predictive control (MPC) \cite{sample_based_mpc} as the local planners that compute the velocity command of the robot. 
Although these modules generally run asynchronously, we process them synchronously for reproducibility.
That is, the time delay of the global planning described in Sec.~\ref{sec:hierarchical_planning} is reproduced as a given constant time: $\Delta t_d = 1$ s.


\subsection{Replanning Strategy Setup}
\honda{
We implemented the rule-based replanning strategies with manually tuned parameters as described in Sec.~\ref{subsec:baselines}, with $d_{\rm{rep}}=1$ m, $\Delta t_{\rm{stuck}}=3$ s, $\Delta t_{\rm{rep}}=1$ s, and $d_{\rm{patience}}=3$ m. We also trained the proposed DRL replanner. The observation space dimension described in Sec.~\ref{subsec:observation} is $62$, with $n_s=20$, $n_p=5$, and $n_t=5$. The DRL replanner was trained using the DQN algorithm described in Sec.~\ref{subsec:training_procedure}, using stable-baselines3~\cite{stable-baselines3}. We used the Adam optimizer to train the Q-network, which features a network architecture consisting of a multi-layer perceptron with hidden layers of $[128, 128]$, a learning rate of $0.0001$, a batch size of $128$, a buffer size of $100k$, and a discount factor $\gamma$ of $0.99$ over $100k$ timesteps (spanning $1k$ episodes). 
}

\subsection{Evaluation Metrics}
To evaluate the effectiveness of our RL-based replanning compared to the rule-based strategies, we simulated 100 trials of the navigation. The following five metrics are used for a quantitative evaluation of the navigation performance:
\begin{itemize}
    \item SR: The success rate over 100 trials, where success is defined as the robot reaching the goal without collision.
    \item CR: The collision rate over 100 trials.
    \item SGT: The SGT in (\ref{eq:sgt}) ($\alpha=4, \beta =8$).
    \item SPL: The average success-weighted normalized path length defined as $\rm{SPL} = \frac{1}{N} \sum^{N}_{i=0}\frac{\myvec{1}_{\rm{suc}}^i \rm{AL}_i}{\max(\rm{AL}_i, \rm{OL}_i)}$
    , where $N$ is the number of trials ($N=100$), $\myvec{1}_{\rm{suc}}^i$ is a binary indicator function of success, and AL and OL denote the actual and optimal traversal length as an indicator of the difficulty of the environment, respectively. The SPL evaluates the navigation's efficiency with respect to the traveled distance.
    \item NR: The number of replanning over 100 trials.
\end{itemize}

\begin{table*}[t]
\centering
\caption{Simulation Results}
\vspace{-3mm}
\label{tab:simulation_result}
\begin{threeparttable}
\begin{tabular}{c|ccccc|ccccc|ccccc}
\toprule
No. of no-entry areas & \multicolumn{5}{c|}{9} & \multicolumn{5}{c|}{16} & \multicolumn{5}{c}{25} \\
Metrics & SR$\uparrow$ & CR$\downarrow$ & SGT$\uparrow$ & SPL$\uparrow$ & NR$\downarrow$ & SR$\uparrow$ & CR$\downarrow$ & SGT$\uparrow$ & SPL$\uparrow$ & NR$\downarrow$ & SR$\uparrow$ & CR$\downarrow$ & SGT$\uparrow$ & SPL$\uparrow$ & NR$\downarrow$ \\ 
\midrule
No replan & 33 & 11 & 0.461 & 0.330 & -- & 27 & 10 & 0.439 & 0.270 & -- & 24 & \textbf{4} & 0.418 & 0.240 & -- \\
\midrule
Distance-based & 66 & 13 & 0.529 & 0.572 & 2202 & 62 & 12 & 0.509 & 0.547 & 2186 & 82 & 12 & 0.561 & 0.705 & 1995 \\
Stuck-based & \textbf{67} & 12 & 0.511 & \textbf{0.619} & 914 & 64 & 13 & 0.493 & 0.605 & 739 & 71 & 6 & 0.482 & 0.658 & 818 \\
Time-based & 64 & 15 & 0.527 & 0.570 & 3063 & 70 & 10 & 0.538 & 0.615 & 3076 & 79 & 12 & 0.562 & 0.688 & 2671 \\
Time w/ patience & 64 & 15 & 0.529 & 0.570 & 2953 & 70 & 10 & 0.540 & 0.615 & 2956 & 79 & 12 & 0.564 & 0.688 & 2558 \\
\midrule
\rowcolor[HTML]{E6E6EF}
DRL replanner (ours) & 64 & \textbf{10} & \textbf{0.532} & 0.567 & 1361 & \textbf{77} & \textbf{4} & \textbf{0.563} & \textbf{0.668} & 2577 & \textbf{87} & 6 & \textbf{0.600} & \textbf{0.751} & 2066 \\ 
\bottomrule
\end{tabular}
\begin{tablenotes}[flushleft]
\item[ ] SR, CR, SGT, SPL, and NR mean success rate [\%], collision rate [\%], success rate weighted goal time, success rate weighted path length, and the number of replanning operations, respectively. SGT and SPL show the time and path efficiency of the navigation. Dijkstra and DWA planners are used for global and local planners in the training and evaluation.
\end{tablenotes}
\end{threeparttable}
\vspace{-5mm}
\end{table*}

\subsection{Experimental Results}
\label{sec:result}
\subsubsection{Quantitative Comparisons}
\label{subsec:comparison_with_baseline}

Table \ref{tab:simulation_result} lists the quantitative evaluation results of 100 trials for each map layout with Dijkstra and DWA planners.
Overall, we confirmed that the baseline strategies, \ie, distance-based, stuck-based, time-based, and time w/ patience, show quite different tendencies for each environment. The stuck-based strategy demonstrates a consistently lower number of replanning operations (NR) and consequently outperforms the other methods in terms of SPL (\ie, shorter travel distance on average) in the easiest environment with a lower number of no-entry areas ($N=9$). For more complicated environments with $N=16, 25$, the distance-based and time-based strategies become more robust and efficient. 
This is arguable because these methods periodically perform replanning to refine the reference path. 
Nevertheless, these results come at the cost of an increase in the number of replanning operations--in other words, more computational resources are required.
In contrast, the DRL replanner, which learns from its experiences to seek better-replanning timings, works comparably well or sometimes substantially better than the other rule-based strategies in each environment. For the $N=9$ environment, the DRL replanner was slightly outperformed by the stuck-based strategy but still able to perform on par with the remaining baselines with much fewer replanning operations.

\subsubsection{Qualitative Results}
\label{subsubsec:qualitative}

Figure \ref{fig:simulation_result} visualizes some selected navigation results with the stuck-based and time-based strategies as well as the DRL replanner. Each method shows its own unique behavior. For example, with the stuck-based strategy in Fig.~\ref{fig:sim_result_16pillar}, the robot simply waited until a dynamic obstacle near the start point was out of the way, and then performed the replanning around the goal by getting stuck at the static obstacle. In contrast, the time-based strategy periodically updated the reference path, but this sometimes resulted in path oscillation, as shown in Fig.~\ref{fig:sim_result_16pillar}. This happens when the frequency of replanning is unnecessarily high compared to the actual need. Figure~\ref{fig:sim_result_25pillar} shows a more challenging case where both stuck-based and time-based replanning failed. In contrast, the DRL replanner was able to learn to adapt its replanning strategy to the given environments and performed replanning only when necessary. In fact, fewer replanning operations were performed in less congested areas, resulting in overall shorter travels compared to the stuck-based and time-based strategies.

\subsubsection{Results with Different Planners}

\begin{table}[t]
\centering
\caption{Comparison of performance by planners}
\vspace{-3mm}
\label{tab:train_other_planners}
\begin{threeparttable}
\begin{tabular}{cccrrrrr}
\toprule
GP & LP & Method & \multicolumn{1}{c}{SR} & \multicolumn{1}{c}{CR} & \multicolumn{1}{c}{SGT} & \multicolumn{1}{c}{SPL} & \multicolumn{1}{c}{NR} \\ \hline
\multirow{3}{*}{Dijkstra} & \multirow{3}{*}{DWA} & Stuck-based & 64 & 13 & 0.493 & 0.605 & 739 \\
 &  & Time-based & 70 & 10 & 0.538 & 0.615 & 3076 \\
 &  & DRL & \textbf{77} & \textbf{4} & \textbf{0.563} & \textbf{0.668} & 2577 \\ \hline
\multirow{3}{*}{Dijkstra} & \multirow{3}{*}{MPC} 
& Stuck-based & 70 & 10 & 0.434 & 0.483 & 298 \\
 &  & Time-based & 75 & 8 & 0.487 & 0.560 & 2181 \\
 &  & DRL & \textbf{82} & \textbf{8} & \textbf{0.502} & \textbf{0.601} & 1582 \\ \hline
\multirow{3}{*}{PRM} & \multirow{3}{*}{DWA} 
& Stuck-based & 66 & 10 & 0.537 & 0.617 & 808 \\
&  & Time-based & 72 & 13 & 0.542 & 0.623 & 2958 \\
&  & DRL & \textbf{74} & \textbf{7} & \textbf{0.570} & \textbf{0.647} & 2293 \\ \hline
 \multirow{3}{*}{RRT*} & \multirow{3}{*}{DWA} 
 & Stuck-based & \textbf{65} & 10 & \textbf{0.473} & \textbf{0.537} & 661 \\
 &  & Time-based & 57 & 6 & 0.446 & 0.441 & 3758 \\
 &  & DRL & 58 & \textbf{5} & 0.445 & 0.443 & 2801 \\ 
 \bottomrule
\end{tabular}
\begin{tablenotes}[flushleft]
\item[ ] GP and LP are global and local planners, respectively.
\end{tablenotes}
\end{threeparttable}
\vspace{-7mm}
\end{table}

Table~\ref{tab:train_other_planners} lists the results of the Dijkstra--MPC, PRM--DWA, and RRT*--DWA combinations of global and local planners, in addition to the Dijkstra--DWA reported in the previous section. We confirmed that the proposed DRL replanner could learn to adapt to the choice of planners, except when RRT* was used. A possible reason for the degraded performance with RRT* is its stochastic nature, which does not give exactly consistent paths when replanning, regardless of the current observation. This would make learning the replanning strategy harder than when combined with the other global planners. Note that PRM can produce a consistent path once the roadmap is generated by sampling, making it a better choice when a sampling-based global planner is required for larger environments.

\subsubsection{Ablation Study}
\label{subsec:training_result}
\begin{table}[t]
\centering
\caption{Comparison of results with RL algorithms}
\vspace{-3mm}
\label{tab:result-with-RL-alg}
\begin{threeparttable}
\begin{tabular}{cccccc}
\toprule
RL algorithm & SR & CR & SGT & SPL & NR \\ \hline
w/o PER & 69 & 12 & 0.514 & 0.595 & 2396 \\
w/ PER-TDerror & 69 & 9 & 0.541 & 0.608 & 2414 \\
w/ PER-Qerror (ours) & \textbf{77} & \textbf{4} & \textbf{0.563} & \textbf{0.668} & 2577 \\ 
\bottomrule
\end{tabular}
\begin{tablenotes}[flushleft]
\item[ ] Dijkstra and DWA planners are used on the 16-pillar map.
\end{tablenotes}
\end{threeparttable}
\vspace{-7mm}
\end{table}

Finally, Table~\ref{tab:result-with-RL-alg} compares other RL techniques that were not used in the proposed method. Specifically, we investigated how the overall performances change if the prioritized experience replay was not used (w/o PER) or if the priority was determined using TD error (w/ PER-TDerror). Although the number of replanning operations (NR) was almost the same, there was a substantial difference in the success rate (SR) and consequently in other metrics such as SGT and SPL. These findings suggest that changes in Q-value can function as a salient clue for replanning, and using them as the priority of experiences leads to more efficient training of the replanning controller.

%% file: src/conclusion.tex
\section{CONCLUSION AND LIMITATIONS}
\label{sec:conclusion}

This paper is the first to delve into when to replan the reference path in a common hierarchical planning framework. We propose the DRL-based solution that addresses inefficiencies caused by the conventional rule-based replanning strategy. Our simulation results demonstrate that the proposed DRL-based replanning strategy achieves similar or better efficiency than the other rule-based strategies in the spaces with branching pathways and dynamic obstacles. We believe that these results have strong implications for designing replanning strategies in autonomous robot navigation. In this study, we used simple \honda{pre-built map} layout to highlight the effect of unforeseen and dynamic obstacles. \honda{Despite the promising results, applying the DRL strategy in real-world scenarios remains challenging due to the diversity of the map layouts and the uncertainty in observations. Future work will aim to enhance our DRL strategy to increase its robustness in real-world, including testing within complex map layouts and improving adaptability to ensure its practical effectiveness.}
